\pdfoutput=1

\documentclass[11pt]{article}

\usepackage{naacl2021}

\usepackage{times}
\usepackage{latexsym}
\usepackage{xspace,amsmath,amsfonts}
\usepackage{bbold}
\usepackage{graphicx}
\usepackage{multirow}
\usepackage{graphicx}

\usepackage[T1]{fontenc}

\usepackage[utf8]{inputenc}

\usepackage{microtype}

\newcommand{\hotpotqa}{HotpotQA\xspace}
\newcommand{\narrativeqa}{NarrativeQA\xspace}
\newcommand{\triviaqa}{TriviaQA\xspace}

\newcommand{\bert}{\textsc{BERT}\xspace}
\newcommand{\roberta}{\textsc{RoBERTa}\xspace}

\newcommand{\etc}{\textsc{ETC}\xspace}
\newcommand{\bigbird}{\textsc{BigBird}\xspace}

\newcommand{\hibert}{\textsc{HIBERT}\xspace}

\newcommand{\rit}{\textsc{ReadTwice}\xspace}
\newcommand{\rite}{\textsc{ReadTwice-E}\xspace}
\newcommand{\ritess}{\textsc{ReadTwice-E(SS)}\xspace}

\title{\rit: Reading Very Large Documents with Memories\vspace*{0.5cm}}

\author{Yury Zemlyanskiy\Thanks{Work is done while at Google} \\
  University of Southern California \\
  \small{\texttt{yury.zemlyanskiy@usc.edu}} \\
  \And
  Joshua Ainslie \\
  Google Research \\
  \small{\texttt{jainslie@google.com}} \\
  \And
  Michiel de Jong \\
  University of Southern California \\
  \small{\texttt{msdejong@usc.edu}} \\
  \AND
  Philip Pham \\
  Google Research \\
  \small{\texttt{phillypham@google.com}} \\
  \And
  Ilya Eckstein \\
  Google Research \\
  \small{\texttt{ilyaeck@google.com}}\\
  \And
  Fei Sha\Thanks{ On leave from University of Southern California (feisha@usc.edu)}    \\
  Google Research \\
  \small{\texttt{fsha@google.com}}}
 
\begin{document}
\maketitle
\begin{abstract}
Knowledge-intensive tasks such as question answering often require assimilating information from different sections of large inputs such as books or article collections. We propose \rit\footnote{Source code and pre-trained checkpoints for \rit can be found at \url{https://goo.gle/research-readtwice}.}, a simple and effective technique that combines several strengths of prior approaches to model long-range dependencies with Transformers. The main idea is to read text in small segments, in parallel, summarizing each segment into a memory table to be used in a second read of the text. We show that the method outperforms models of comparable size on several question answering (QA) datasets and sets a new state of the art on the challenging \narrativeqa task, with questions about entire books. 
\end{abstract}

\begin{figure*}[t]
    \centering
    \includegraphics[width=0.85\textwidth]{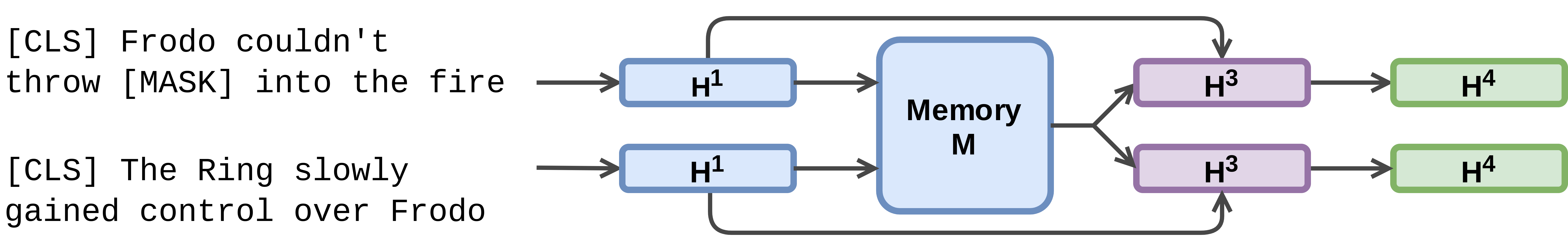} 
    \caption{\rit model architecture. The input is processed twice, with a memory table for inter-segment information sharing.}
    \label{fig:readittwice}
\end{figure*}

\section{Introduction}
\label{sec:intro}

Transformer-based models such as BERT are very effective in capturing long-range dependencies in text passages through the attention mechanism~\citep{transformer,bert}. However, the amount of compute in  attention depends quadratically on the number of tokens in an input text passage. As such, the standard BERT implementation limits input size to a fixed number (often 512) of tokens.

In reality, dependencies over significantly longer ranges are common and modeling them is crucial. For instance, in a sentence like \emph{Inside the Sammath Naur, the Ring-bearer struggled to throw the Ring into the volcano},  the narrative interweaves several prior storylines from a book. Comprehending this sentence therefore requires looking up previous mentions of \emph{Ring-bearer} and \emph{Sammath Naur}, located many tokens away. 

Several methods have been proposed to address this challenge; see ~\citep{transformer_survey} for a survey and \S\ref{sec:related_Work} for a detailed discussion. One popular strategy is to reduce the number of tokens attended to. Longer inputs can in fact be processed in this way -- but only up to a limit of around 5,000 tokens, as used in ~\citep{etc,bigbird,longformer} -- far below the context sizes required to model long documents such as books. 

Another strategy such as \hibert~\citep{hibert} splits inputs into smaller segments which are processed individually, then assembled into a hierarchical representation. As a downside, inter-segment context is unavailable during encoding.

We propose \rit, a simple approach that combines the strengths of both strategies. As its name suggests, the main idea is to process the input twice: a long text input (such as a document, or even a book) is treated as a collection of shorter text segments which are read independently and in parallel. Then, the encoder reads each segement again, now \emph{augmented} with compressed information from other segments. 

The crucial component in \rit, as illustrated in Figure~\ref{fig:readittwice}, is a memory module that holds compressed information from all segments. That compressed information is used only \emph{once}: in the second pass. Thus, \rit is much more computationally efficient than models like \etc that rely on memory for all segments, in every layer. While \rit requires two passes, it differs from hierarchical models such as \hibert that do not condition segment encoding on other segments. \S\ref{sec:related_Work} contrasts these approaches in more detail.

We validate the efficacy of \rit on extractive question answering (QA) tasks, showing strong performance on \hotpotqa~\citep{hotpotqa}, \triviaqa~\citep{triviaqa} and \narrativeqa~\citep{narrativeqa}. In particular, \rit significantly improves the state-of-the-art on QA based on \textit{entire books} in \narrativeqa, with absolutes gains of 4.5 ROUGE-L points and 3 BLEU-1 points (relative improvements of 23\% and 17\%, respectively).

\section{Method}
\label{sec:method}

We first describe the \rit model, followed by its pre-training procedure.
\subsection{\rit}

The model reads a large text document split into $N$ segments $x_1, \ldots, x_N$; each $x_i$ is limited to 512 tokens, as in a typical BERT model. 

The model architecture is depicted in Figure~\ref{fig:readittwice}. In the first read, each segment is encoded independently with  standard BERT. Then, memories are extracted from each segment---a process we describe in detail later---and gathered into a global memory pool. For the second read, a $\mathtt{MemoryAttention}$ layer (with a residual connection and a $\mathtt{LayerNorm}$ on top) is first used to merge the information from the former intra-segmental contextual token embeddings and the global memory. The merged result is then read by another small BERT model with only \textit{two} Transformer layers to produce the final output. The rationale is that the first read already generates rich contextualized embeddings, and the second read only needs to incorporate information from the memory. More formally:
\begin{align*}
    H^0_i &= \mathtt{TokenEmbed}(x_i), H^1_i = \mathtt{BERT}_1(x_i),  \forall\ i\\
    M_i &= \mathtt{ExtractMemories}(H^1_i), \forall i \\
    M & = \mathtt{Gather}([M_1, \ldots, M_N])\\
    H^2_i &= \mathtt{MemoryAttention}(H^1_i, M), \forall i\\
    H^3_i & = \mathtt{LayerNorm}(H^1_i + H^2_i), \forall\ i\\
    H^4_i &= \mathtt{BERT}_2(H^3_i), \forall\ i
\end{align*}
Next, we describe the newly introduced layers.

\paragraph{$\mathtt{ExtractMemories}$ \textmd{and} $\mathtt{Gather}$} Our aim is to compress the information in each segment and disseminate it to other segments to be used in the second read. We consider three types of memories:
\begin{itemize}
    \item \rit($\mathtt{CLS}$). One obvious choice is to use the \textsc{cls} token representation associated with segment $x_i$ as a summary of the segment.
    
    \item \rit($\mathtt{STS}$). To obtain more fine-grained memories, we extract a memory vector for each consecutive span of 32 tokens. Contextual embeddings of each span's first and the last tokens are concatenated and linearly projected to a single point in the token vector space as the span representation. The projection matrix is learned end to end.     

    \item \rit($\mathtt{E}$). In another variant of span-based memory, we memorize representations of entity mention spans. To obtain these spans, we first annotate each segment with an external Named Entity Recognition system. Then, each entity mention span is encoded in the same way as in \rit($\mathtt{STS}$). This design is motivated by the intuition that long-range dependencies primarily occur between entities. 
\end{itemize}
Empirically, we find that \rit($\mathtt{E}$) leads to best performance (see the ablation in Section \ref{sec:exp:ablation}) and it is the memory type used in our headline results. 

We collect all memories from all segments into a flat memory table. The table size is given by the number of segments ($\mathtt{CLS}$), the number of 32-token spans ($\mathtt{STS}$), or the number of entity mentions ($\mathtt{E}$). 

\paragraph{$\mathtt{MemoryAttention}$} In this layer, we let contextual token embeddings from individual segments interact with other segments' memories via dot-product attention over the memory table. 

Let $h_{ij}$ be the contextual embedding of token $j$ in segment $i$ after the first read. And let $m$ be a memory table entry whose source segment is given by $m_s$. We then define its attention weight as:
\begin{align}
    \alpha_m = \frac{e^{h_{ij}^TM_m  + r_{i,m_s}}}{\sum_m e^{h_{ij}^TM_m + r_{i,m_s}} + e^{h_{ij}^T M_0}}
\end{align}
where $M_0$ is a learnable no-op memory not associated with any specific text. $r_{i, m_s}$ is a learned position score which captures the relative distance between segment $i$ and the memory $M_m$, akin to \citet{relattn}:
\begin{equation}
    r_{i, m_s} = \omega( \mathtt{dist}(i, m_s))
\end{equation}
where $\omega$ is a set of weights indexed by the distance
\begin{equation}
     \mathtt{dist}(i, m_s) = \left\{\begin{array}{ll}
     -B &  i-m_s < -B\\
     B & i-m_s > B\\
     i-m_s &\mathtt{otherwise}
     \end{array} \right.
\end{equation}
where the cutoff threshold $B$ clips the effect of distance to $[-B, B]$. We set $B$ to 10 in this work.

Finally, the $\mathtt{MemoryAttention}$ layer output for a given token is given by
\begin{align}
    h_{ij}^2 = \sum_{m=1} \alpha_m M_m
\end{align}

\subsection{Pre-training}

We pretrain \rit similarly to \citep{bert}, using the Wikipedia and BooksCorpus datasets. When entity mentions are used in the memory table, the texts are processed with the Entity Linking (EL) and Named Entity Recognition (NER) tools from the Google Cloud NLP API\footnote{\url{https://cloud.google.com/natural-language/docs/basics\#entity_analysis}}. Moreover, we use existing hyperlinks in Wikipedia as additional entity annotations. The first and the second BERT readers are trained end-to-end.

Our pre-training objective is the standard Masked Language Model (MLM) task, with the MLM prediction loss computed based on the output of the second reader. 

In order to encourage the model to rely on the memory, we increase the difficulty of the MLM task. Following the entity masking procedure in \citep{realm,earnie}, we mask entity mention tokens more aggressively at a 25\% rate and jointly mask all tokens within a mention. By contrast, for non-entity tokens, we mask contiguous sequences of random length at a 15\% rate.
\section{Related Work}
\label{sec:related_Work}
One way to extend the limit on input size is by reducing the number of tokens attended to.
\textsc{etc}~\citep{etc} and \textsc{longformer}~\citep{longformer} allow standard attention only between tokens within a fixed distance. To allow information flow over longer distances, they use auxiliary global "memory" tokens which attend to all regular tokens and vice versa. \bigbird~\citep{bigbird} additionally has each token attend to a random subset of other tokens. While reducing asymptotic complexity from quadratic to linear (in input size), these global tokens are added at each attention layer, incurring a high computational cost.

Another approach is to split the input into multiple segments and then aggregate information across segments. This is achieved through hierarchical modeling ~\citep{helmo,hibert}. While reducing the attention size to the number of segments, each individual segment has no information about its siblings during token-level encoding. Alternatively, recurrent models \citep{dai2019transformer,rae2019compressive} read a large input from left to right, dynamically compressing faraway contexts, thus allowing unidirectional information aggregation (left to right). One disadvantage is that the input needs to be processed sequentially, which becomes time-consuming for producing contextualized representations of a large input.

Our method brings these lines of work together. Processing segments independently and in parallel, then memorizing their compressed representations and sharing memory across segments enables contextual embeddings to be updated based on faraway information. Enabling memory sharing only once---during the second read---allows it be done cheaply. 

Note that the memory module here is internally generated from the input, as opposed to external memory models which are orthogonal to our approach~\citep{knowbert,fevry2020eae}.
\section{Experiments}
\label{sec:exp}

\subsection{Pre-training setup}
All \rit models are initialized with the public \roberta (base) checkpoint\footnote{\url{https://dl.fbaipublicfiles.com/fairseq/models/roberta.base.tar.gz}} adapted to Tensorflow by \citet{tf_roberta}. Further, models are pre-trained for 1M steps on 64 TPU cores using the LAMB optimizer \citep{lamb}.

Each batch contains 512 segments, with at most 128 segments per document. The segments are consecutive spans of 512 tokens. Therefore, the model can process documents up to 65k ($\approx 128 \times 512$) tokens. Each batch contains the maximum number of documents such that the total number of segments is at most 512. Approximately half of Wikipedia articles fit in one segment (thus not needing memory), with a fat tail of longer documents.

In terms of compute and memory overhead, \rit is about 30\% slower than the \roberta-base model and uses 15M (or 12\%) more parameters: 14M owing to the second read $\mathtt{BERT}_2$ and 1M due to $\mathtt{ExtractMemories}$ and $\mathtt{MemoryAttention}$ layers.

\subsection{Evaluation setup}
We evaluate \rit on the downstream extractive question-answering task using several datasets: \hotpotqa (HQA)~\citep{hotpotqa}, \triviaqa (TQA)~\citep{triviaqa} and \narrativeqa (NQA)~\citep{narrativeqa}.

In HQA, questions are based on relatively short text passages (2 evidence paragraphs), with eight additional distractor passages. In TQA, evidence text is medium-sized. NQA asks questions about entire books, requiring a successful QA system to model very long-range dependencies. The NQA dataset has an average of 62,000 words per document with a maximum of 400,000. Only 40\% of NQA's answers are span-based --  we use a ROUGE-L oracle as training labels for the other questions.

\rit is fine-tuned on each task. QA-specific heads are used to generate span-based predictions, consisting of fully-connected layers that take contextual embeddings from the second reader as inputs. These layers output a score for whether the corresponding tokens are the beginning or ending of an answer span. For a similar setup, see multi-segment based QA tasks~\citep{ClarkGardner,ProbAssumptionsQA}.

During fine-tuning, batches contain 128 segments for all tasks (also with up to 128 segments per document). Every segment contains 512 tokens, but as neighboring segments have 128 token overlaps, the model can process documents of up to 49K tokens ($\approx 128 \times (512-128)$). For TQA and HQA, documents have approximately 10 segments. For NQA, we split the documents into sub-documents with 49k tokens and apply memory only within these sub-documents.

We perform hyperparameter search only over learning rate  $\lambda \in \{5e-6,\ 1e-5,\ 3e-5\}$ and train for 6 epochs with 10\% warm up proportion. Moreover, we use early stopping based on the performance on the development set.

\begin{table}
\centering
\begin{tabular}{p{2.3cm}r|rr}
    \multirow{2}{*}{\textbf{Model}} & \textbf{\textsc{HQA}} & \multicolumn{2}{c}{\textbf{\textsc{TQA}}} \\
    &   \textbf{F1 (ans)} & \textbf{F1(dev)} & \textbf{F1(test)}\\
    \hline
    LF & 74.3 & 75.2 &- \\
    \etc &  75.1 & -&-\\
    \bigbird & 75.7 & 79.5 & -\\
    \hline
    \mbox{\roberta (us)} &  72.0 & 75.9 &-  \\
    \mbox{\rite}  & \textbf{75.9} & \textbf{80.7} & \textbf{80.9}\\ 
\end{tabular}
\caption[This text is not visible and used to put a footnote]{\label{table:hotpot_trivia_qa} Results on HotpotQA development set (\textbf{answer only} F1 score) and on TriviaQA development and test splits for the Wikipedia full setting. Additional test results are available on the public leaderboard\protect\footnotemark}
\end{table}

\footnotetext{See \url{https://competitions.codalab.org/competitions/17208\#results}, tab ``Wikipedia''.}

\begin{table*}[t]
\centering
\begin{tabular}{lrrrrr}
    \textbf{Model} & \textbf{ROUGE-L} & \textbf{BLEU-1} & \textbf{BLEU-4} & \textbf{METEOR}\\
    \hline
    BiDAF \citep{narrativeqa} & 6.3 / 6.2 & 5.8 / 5.7 & 0.2 / 0.3 & 3.8 / 3.7 \\
    $R^3$ \citep{r3}  & 11.4 / 11.9 & 16.4 / 15.7 & 0.5 / 0.5 & 3.5 / 3.5 \\
    BM25+BERT \citep{frhard} & 14.8 / 15.5 & 14.6 / 14.5 & 1.8 / 1.4 & 5.1 / 5.0 \\
    \hline
    \roberta (us) & 17.4 / 18.0 & 18.2 / 18.0 & 2.4 / 2.6 & 5.4 / 5.4 \\
    \etc (us) & 18.3 / 18.8 & 16.1 / 17.2 & 2.4 / 2.7 & 5.4 / 5.4\\
    \hline
    \rit($\mathtt{E}$) & \textbf{22.7} / \textbf{23.3} & \textbf{21.1} / \textbf{21.1} & \textbf{3.6} / \textbf{4.0} & \textbf{6.7} / \textbf{7.0} \\    
\end{tabular}
\caption{Results on the NarrativeQA's development / test splits.
}
\label{table:narrative_qa}
\end{table*}

\subsection{Main Results}

Results for HQA and TQA are reported in Table~\ref{table:hotpot_trivia_qa}. We compare to prior art (using reported results where available or from our own implementations otherwise, denoted as ``us''): Longformer (LF)~\citep{longformer}, ETC~\citep{etc}, BigBird~\citep{bigbird}, and  \roberta~\citep{roberta}. By default, we compare against the ``base`` configuration of those models where the number of parameters is comparable to \bert-Base, as is the case for \rit.

Table~\ref{table:hotpot_trivia_qa} shows that for small to medium sized text passages, the proposed \rit outperforms all models of comparable size.

Table~\ref{table:narrative_qa} contrasts \rit to other methods on extremely large contexts: BiDAF~\citep{narrativeqa}, $R^3$ \citep{r3},  BM25 + BERT Reader / Ranker~\citep{frhard} and our own implementation of \roberta and \etc\footnote{For ETC we use the public (base configuration) checkpoint \url{https://storage.googleapis.com/gresearch/etcmodel/checkpoints/etc_base_2x_pretrain.zip}}. 
\rit significantly outperforms all previous work and establishes new state-of-the-art results, demonstrating the effectiveness of performing a second read conditioned on global memory for processing extremely long texts.  

\subsection{Ablation Analysis \& Discussion}
\label{sec:exp:ablation}

To isolate individual components' contributions, Table~\ref{table:ablations} contrasts several variants of \rit. These ablations lead to two key insights.

\paragraph{Inter-segment memory matters} We introduce a variant \ritess (where SS stands for ``Single Segment'') to isolate the gains from the memory layer. \ritess prevents segments from attending to memories of other segments, thus disabling long-range dependency modeling. We observe that \rite improves over \ritess on all tasks, modestly but non-negligibly for TQA, and significantly for HQA and especially NQA. 

This matches our knowledge of those datasets: TQA questions are based on a relatively short context and can typically be answered using a single passage in the context document. HQA questions have a similarly sized context, but are explicitly constructed to require information from multiple paragraphs to answer, and \rit shows accordingly larger gains. Finally, NQA has much larger contexts, and its questions generally require information from different parts of the document, increasing the importance of long-range dependency modeling and accordingly, the performance boost from \rit.

\paragraph{Entities matter}
Entity mentions appears to be the most effective memory type in most experiments, leading to noticeably improved performance on both HQA and NQA. The difference is most pronounced in NQA whose particularly long and challenging contexts make it a perfect testbed.

\paragraph{Source of non-memory gains}
The non-memory gains over a baseline \roberta model originate from the two extra layers and the entity-based MLM objective. In order to disentangle the sources of gains we train the \ritess model using a 10-layer Transformer for $\mathtt{BERT}_1$ (denoted as $\mathtt{E(SS,10L)}$ in Table~\ref{table:ablations}), with the same number of layers as \roberta. While the gains from 2 extra layers are significant ($\mathtt{E(SS)}$ vs $\mathtt{E(SS,10L)}$), most of the gains appear to result from the custom pre-training procedure ($\mathtt{E(SS,10L)}$ vs \roberta).

\begin{table}[t]
\centering
\begin{tabular}{p{1.5cm}p{0.9cm}p{1.3cm}p{1.3cm}p{0.8cm}}
    \textbf{Model} & \textbf{\textsc{HQA}} & \textbf{\textsc{NQA}}-R & \textbf{\textsc{NQA}}-B & \textbf{\textsc{TQA}}\\
    \hline
    $\mathtt{E}$ & 75.89 & 22.71 & 21.07 & 80.7\\
    \hline
    $\mathtt{E(SS)}$ & 75.08 & 21.93 & 18.39 & 80.3\\
    \hline
    $\mathtt{E(SS, 10L)}$ & 74.70 & 21.39 & 18.37 & 80.4\\ 
    $\roberta$ & 72.00 & 17.40 & 18.2 & 75.9 \\
    \hline
    $\mathtt{CLS}$ & 75.32 & 20.89 & 17.80 & 80.6\\    
    $\mathtt{STS}$  & 75.39 & 21.08 & 18.38 & 80.4\\    
    \hline
\end{tabular}
\caption{\label{table:ablations} Ablation studies on variants of \rit on the dev sets. We report F1 (answer only) score for \textsc{HQA}, ROUGE-L and BLEU-1 for \textsc{NQA} (denoted -R and -B respectively) and F1 for \textsc{TQA}.}
\end{table}

\section{Conclusion \& Future Work}
\label{sec:conclusion}

\rit performs well on several QA tasks, particularly NarrativeQA where long-range dependencies among entities appear to be very important. The proposed method is conceptually simple, easy to implement and is capable of reading entire books. For future work, we plan to explore new memory types, hierarchies and aggregation functions. We also aim to apply the model to other tasks, particularly long text summarization, likely to benefit from a memory-forming mechanism.

\section*{Acknowlegements}
We thank Santiago Ontanon, Manzil Zaheer, Sudeep Gandhe, Anirudh Ravula, Bhargav Kanagal, Jules Gagnon-Marchand, Sumit Sanghai, Ming-Wei Chang and Kenton Lee for insightful discussions, \citet{frhard} for a sample evaluation code for \narrativeqa and reviewers for their feedback.
This work is partially supported by NSF Awards IIS-1513966/ 1632803/1833137, CCF-1139148, DARPA Awards\#: FA8750-18-2-0117, FA8750-19-1-0504,  DARPA-D3M - Award UCB-00009528, Google Research Awards, gifts from Facebook and Netflix, and ARO\# W911NF-12-1-0241 and W911NF-15-1-0484.


\begin{thebibliography}{26}
\expandafter\ifx\csname natexlab\endcsname\relax\def\natexlab#1{#1}\fi

\bibitem[{Ainslie et~al.(2020)Ainslie, Ontanon, Alberti, Cvicek, Fisher, Pham,
  Ravula, Sanghai, Wang, and Yang}]{etc}
Joshua Ainslie, Santiago Ontanon, Chris Alberti, Vaclav Cvicek, Zachary Fisher,
  Philip Pham, Anirudh Ravula, Sumit Sanghai, Qifan Wang, and Li~Yang. 2020.
\newblock \href {https://www.aclweb.org/anthology/2020.emnlp-main.19} {{ETC}:
  Encoding long and structured inputs in transformers}.
\newblock In \emph{Proceedings of the 2020 Conference on Empirical Methods in
  Natural Language Processing (EMNLP)}, pages 268--284, Online. Association for
  Computational Linguistics.

\bibitem[{Beltagy et~al.(2020)Beltagy, Peters, and Cohan}]{longformer}
Iz~Beltagy, Matthew~E. Peters, and Arman Cohan. 2020.
\newblock \href {http://arxiv.org/abs/2004.05150} {Longformer: The
  long-document transformer}.
\newblock \emph{CoRR}, abs/2004.05150.

\bibitem[{Chang et~al.(2019)Chang, Toutanova, Lee, and Devlin}]{helmo}
Ming{-}Wei Chang, Kristina Toutanova, Kenton Lee, and Jacob Devlin. 2019.
\newblock \href {http://arxiv.org/abs/1901.09128} {Language model pre-training
  for hierarchical document representations}.
\newblock \emph{CoRR}, abs/1901.09128.

\bibitem[{Cheng et~al.(2020)Cheng, Chang, Lee, and
  Toutanova}]{ProbAssumptionsQA}
Hao Cheng, Ming{-}Wei Chang, Kenton Lee, and Kristina Toutanova. 2020.
\newblock \href {https://www.aclweb.org/anthology/2020.acl-main.501/}
  {Probabilistic assumptions matter: Improved models for distantly-supervised
  document-level question answering}.
\newblock In \emph{Proceedings of the 58th Annual Meeting of the Association
  for Computational Linguistics, {ACL} 2020, Online, July 5-10, 2020}, pages
  5657--5667. Association for Computational Linguistics.

\bibitem[{Clark and Gardner(2018)}]{ClarkGardner}
Christopher Clark and Matt Gardner. 2018.
\newblock \href {https://doi.org/10.18653/v1/P18-1078} {Simple and effective
  multi-paragraph reading comprehension}.
\newblock In \emph{Proceedings of the 56th Annual Meeting of the Association
  for Computational Linguistics, {ACL} 2018, Melbourne, Australia, July 15-20,
  2018, Volume 1: Long Papers}, pages 845--855. Association for Computational
  Linguistics.

\bibitem[{Dai et~al.(2019)Dai, Yang, Yang, Carbonell, Le, and
  Salakhutdinov}]{dai2019transformer}
Zihang Dai, Zhilin Yang, Yiming Yang, Jaime Carbonell, Quoc~V Le, and Ruslan
  Salakhutdinov. 2019.
\newblock Transformer-xl: Attentive language models beyond a fixed-length
  context.
\newblock \emph{arXiv preprint arXiv:1901.02860}.

\bibitem[{Devlin et~al.(2019)Devlin, Chang, Lee, and Toutanova}]{bert}
Jacob Devlin, Ming{-}Wei Chang, Kenton Lee, and Kristina Toutanova. 2019.
\newblock \href {https://doi.org/10.18653/v1/n19-1423} {{BERT:} pre-training of
  deep bidirectional transformers for language understanding}.
\newblock In \emph{Proceedings of the 2019 Conference of the North American
  Chapter of the Association for Computational Linguistics: Human Language
  Technologies, {NAACL-HLT} 2019, Minneapolis, MN, USA, June 2-7, 2019, Volume
  1 (Long and Short Papers)}, pages 4171--4186. Association for Computational
  Linguistics.

\bibitem[{Févry et~al.(2020)Févry, Soares, FitzGerald, Choi, and
  Kwiatkowski}]{fevry2020eae}
Thibault Févry, Livio~Baldini Soares, Nicholas~Arthur FitzGerald, Eunsol Choi,
  and Tom Kwiatkowski. 2020.
\newblock Entities as experts: Sparse memory access with entity supervision.
\newblock In \emph{EMNLP 2020 - Conference on Empirical Methods in Natural
  Language Processing}.

\bibitem[{Guu et~al.(2020)Guu, Lee, Tung, Pasupat, and Chang}]{realm}
Kelvin Guu, Kenton Lee, Zora Tung, Panupong Pasupat, and Ming{-}Wei Chang.
  2020.
\newblock \href {http://arxiv.org/abs/2002.08909} {{REALM:} retrieval-augmented
  language model pre-training}.
\newblock \emph{CoRR}, abs/2002.08909.

\bibitem[{Joshi et~al.(2017)Joshi, Choi, Weld, and Zettlemoyer}]{triviaqa}
Mandar Joshi, Eunsol Choi, Daniel~S. Weld, and Luke Zettlemoyer. 2017.
\newblock \href {https://doi.org/10.18653/v1/P17-1147} {Triviaqa: {A} large
  scale distantly supervised challenge dataset for reading comprehension}.
\newblock In \emph{Proceedings of the 55th Annual Meeting of the Association
  for Computational Linguistics, {ACL} 2017, Vancouver, Canada, July 30 -
  August 4, Volume 1: Long Papers}, pages 1601--1611. Association for
  Computational Linguistics.

\bibitem[{Kocisk{\'{y}} et~al.(2018)Kocisk{\'{y}}, Schwarz, Blunsom, Dyer,
  Hermann, Melis, and Grefenstette}]{narrativeqa}
Tom{\'{a}}s Kocisk{\'{y}}, Jonathan Schwarz, Phil Blunsom, Chris Dyer,
  Karl~Moritz Hermann, G{\'{a}}bor Melis, and Edward Grefenstette. 2018.
\newblock \href {https://transacl.org/ojs/index.php/tacl/article/view/1197}
  {The narrativeqa reading comprehension challenge}.
\newblock \emph{Trans. Assoc. Comput. Linguistics}, 6:317--328.

\bibitem[{Liu et~al.(2019)Liu, Ott, Goyal, Du, Joshi, Chen, Levy, Lewis,
  Zettlemoyer, and Stoyanov}]{roberta}
Yinhan Liu, Myle Ott, Naman Goyal, Jingfei Du, Mandar Joshi, Danqi Chen, Omer
  Levy, Mike Lewis, Luke Zettlemoyer, and Veselin Stoyanov. 2019.
\newblock \href {http://arxiv.org/abs/1907.11692} {Roberta: {A} robustly
  optimized {BERT} pretraining approach}.
\newblock \emph{CoRR}, abs/1907.11692.

\bibitem[{Mou et~al.(2020)Mou, Yu, Yao, Yang, Guo, Potdar, and Su}]{frhard}
Xiangyang Mou, Mo~Yu, Bingsheng Yao, Chenghao Yang, Xiaoxiao Guo, Saloni
  Potdar, and Hui Su. 2020.
\newblock \href {http://arxiv.org/abs/2007.09878} {Frustratingly hard evidence
  retrieval for {QA} over books}.
\newblock \emph{CoRR}, abs/2007.09878.

\bibitem[{Peters et~al.(2019)Peters, Neumann, IV, Schwartz, Joshi, Singh, and
  Smith}]{knowbert}
Matthew~E. Peters, Mark Neumann, Robert L.~Logan IV, Roy Schwartz, Vidur Joshi,
  Sameer Singh, and Noah~A. Smith. 2019.
\newblock \href {https://doi.org/10.18653/v1/D19-1005} {Knowledge enhanced
  contextual word representations}.
\newblock In \emph{Proceedings of the 2019 Conference on Empirical Methods in
  Natural Language Processing and the 9th International Joint Conference on
  Natural Language Processing, {EMNLP-IJCNLP} 2019, Hong Kong, China, November
  3-7, 2019}, pages 43--54. Association for Computational Linguistics.

\bibitem[{Rae et~al.(2019)Rae, Potapenko, Jayakumar, and
  Lillicrap}]{rae2019compressive}
Jack~W Rae, Anna Potapenko, Siddhant~M Jayakumar, and Timothy~P Lillicrap.
  2019.
\newblock Compressive transformers for long-range sequence modelling.
\newblock \emph{arXiv preprint arXiv:1911.05507}.

\bibitem[{Rothe et~al.(2020)Rothe, Narayan, and Severyn}]{tf_roberta}
Sascha Rothe, Shashi Narayan, and Aliaksei Severyn. 2020.
\newblock \href {https://transacl.org/ojs/index.php/tacl/article/view/1849}
  {Leveraging pre-trained checkpoints for sequence generation tasks}.
\newblock \emph{Trans. Assoc. Comput. Linguistics}, 8:264--280.

\bibitem[{Sharma et~al.(2017)Sharma, Asri, Schulz, and Zumer}]{nlgeval}
Shikhar Sharma, Layla~El Asri, Hannes Schulz, and Jeremie Zumer. 2017.
\newblock \href {http://arxiv.org/abs/1706.09799} {Relevance of unsupervised
  metrics in task-oriented dialogue for evaluating natural language
  generation}.
\newblock \emph{CoRR}, abs/1706.09799.

\bibitem[{Shaw et~al.(2018)Shaw, Uszkoreit, and Vaswani}]{relattn}
Peter Shaw, Jakob Uszkoreit, and Ashish Vaswani. 2018.
\newblock \href {https://doi.org/10.18653/v1/n18-2074} {Self-attention with
  relative position representations}.
\newblock In \emph{Proceedings of the 2018 Conference of the North American
  Chapter of the Association for Computational Linguistics: Human Language
  Technologies, NAACL-HLT, New Orleans, Louisiana, USA, June 1-6, 2018, Volume
  2 (Short Papers)}, pages 464--468. Association for Computational Linguistics.

\bibitem[{Sun et~al.(2019)Sun, Wang, Li, Feng, Chen, Zhang, Tian, Zhu, Tian,
  and Wu}]{earnie}
Yu~Sun, Shuohuan Wang, Yu{-}Kun Li, Shikun Feng, Xuyi Chen, Han Zhang, Xin
  Tian, Danxiang Zhu, Hao Tian, and Hua Wu. 2019.
\newblock \href {http://arxiv.org/abs/1904.09223} {{ERNIE:} enhanced
  representation through knowledge integration}.
\newblock \emph{CoRR}, abs/1904.09223.

\bibitem[{Tay et~al.(2020)Tay, Dehghani, Bahri, and
  Metzler}]{transformer_survey}
Yi~Tay, Mostafa Dehghani, Dara Bahri, and Donald Metzler. 2020.
\newblock \href {http://arxiv.org/abs/2009.06732} {Efficient transformers: {A}
  survey}.
\newblock \emph{CoRR}, abs/2009.06732.

\bibitem[{Vaswani et~al.(2017)Vaswani, Shazeer, Parmar, Uszkoreit, Jones,
  Gomez, Kaiser, and Polosukhin}]{transformer}
Ashish Vaswani, Noam Shazeer, Niki Parmar, Jakob Uszkoreit, Llion Jones,
  Aidan~N. Gomez, Lukasz Kaiser, and Illia Polosukhin. 2017.
\newblock \href {http://papers.nips.cc/paper/7181-attention-is-all-you-need}
  {Attention is all you need}.
\newblock In \emph{Advances in Neural Information Processing Systems 30: Annual
  Conference on Neural Information Processing Systems 2017, 4-9 December 2017,
  Long Beach, CA, {USA}}, pages 5998--6008.

\bibitem[{Wang et~al.(2018)Wang, Yu, Guo, Wang, Klinger, Zhang, Chang, Tesauro,
  Zhou, and Jiang}]{r3}
Shuohang Wang, Mo~Yu, Xiaoxiao Guo, Zhiguo Wang, Tim Klinger, Wei Zhang, Shiyu
  Chang, Gerry Tesauro, Bowen Zhou, and Jing Jiang. 2018.
\newblock \href
  {https://www.aaai.org/ocs/index.php/AAAI/AAAI18/paper/view/16712}
  {R\({}^{\mbox{3}}\): Reinforced ranker-reader for open-domain question
  answering}.
\newblock In \emph{Proceedings of the Thirty-Second {AAAI} Conference on
  Artificial Intelligence, (AAAI-18), the 30th innovative Applications of
  Artificial Intelligence (IAAI-18), and the 8th {AAAI} Symposium on
  Educational Advances in Artificial Intelligence (EAAI-18), New Orleans,
  Louisiana, USA, February 2-7, 2018}, pages 5981--5988. {AAAI} Press.

\bibitem[{Yang et~al.(2018)Yang, Qi, Zhang, Bengio, Cohen, Salakhutdinov, and
  Manning}]{hotpotqa}
Zhilin Yang, Peng Qi, Saizheng Zhang, Yoshua Bengio, William~W. Cohen, Ruslan
  Salakhutdinov, and Christopher~D. Manning. 2018.
\newblock \href {https://doi.org/10.18653/v1/d18-1259} {Hotpotqa: {A} dataset
  for diverse, explainable multi-hop question answering}.
\newblock In \emph{Proceedings of the 2018 Conference on Empirical Methods in
  Natural Language Processing, Brussels, Belgium, October 31 - November 4,
  2018}, pages 2369--2380. Association for Computational Linguistics.

\bibitem[{You et~al.(2020)You, Li, Reddi, Hseu, Kumar, Bhojanapalli, Song,
  Demmel, Keutzer, and Hsieh}]{lamb}
Yang You, Jing Li, Sashank~J. Reddi, Jonathan Hseu, Sanjiv Kumar, Srinadh
  Bhojanapalli, Xiaodan Song, James Demmel, Kurt Keutzer, and Cho{-}Jui Hsieh.
  2020.
\newblock \href {https://openreview.net/forum?id=Syx4wnEtvH} {Large batch
  optimization for deep learning: Training {BERT} in 76 minutes}.
\newblock In \emph{8th International Conference on Learning Representations,
  {ICLR} 2020, Addis Ababa, Ethiopia, April 26-30, 2020}. OpenReview.net.

\bibitem[{Zaheer et~al.(2020)Zaheer, Guruganesh, Dubey, Ainslie, Alberti,
  Onta{\~{n}}{\'{o}}n, Pham, Ravula, Wang, Yang, and Ahmed}]{bigbird}
Manzil Zaheer, Guru Guruganesh, Avinava Dubey, Joshua Ainslie, Chris Alberti,
  Santiago Onta{\~{n}}{\'{o}}n, Philip Pham, Anirudh Ravula, Qifan Wang,
  Li~Yang, and Amr Ahmed. 2020.
\newblock \href {http://arxiv.org/abs/2007.14062} {Big bird: Transformers for
  longer sequences}.
\newblock \emph{CoRR}, abs/2007.14062.

\bibitem[{Zhang et~al.(2019)Zhang, Wei, and Zhou}]{hibert}
Xingxing Zhang, Furu Wei, and Ming Zhou. 2019.
\newblock \href {http://arxiv.org/abs/1905.06566} {{HIBERT:} document level
  pre-training of hierarchical bidirectional transformers for document
  summarization}.
\newblock \emph{CoRR}, abs/1905.06566.

\end{thebibliography}

\clearpage
\appendix
\label{sec:appendix}

\section{Method}
\paragraph{Sparse $\mathtt{MemoryAttention}$}
In the standard setting \rit($\mathtt{CLS}$) and \rit($\mathtt{STS}$) apply the $\mathtt{MemoryAttention}$ layer for all tokens $h_{ij}$. Our preliminary experiments showed that \rit($\mathtt{E}$) benefits from a sparse pattern -- when the layer is applied only over tokens that belong to an entity mention. It acts like an identity function for all other tokens. This follows an intuition that long-range dependencies in text mainly occur between entity mentions. Similar sparse patterns for \rit($\mathtt{CLS}$) and \rit($\mathtt{STS}$) (e.g. CLS tokens attending only CLS-based memories) affected their performance negatively.

\section{Pre-training details}

\paragraph{Entity mention specific pre-training}  We use \textit{coreference resolution} as an auxiliary pre-training task specifically for \rit($\mathtt{E}$). If there are two entities in the memory table pointing to the same entity (but in different segments), our co-reference resolution task will encourage their entity representations to be close. This is achieved through a binary classification task on whether $m$ and $m'$ entries in $M$ point to the same entity. The classification probability is modeled as:
\begin{equation*}
    y_{mm'} = \sigma(M_m^TM_{m'} + b_0) 
\end{equation*}
where $b_0$ is a bias term. A logistic loss is formed (using the ground-truth as positive examples and all other entities as negative examples) and added to the MLM learning objective, for every entry in the memory table. Memories that correspond to mentions without a corresponding entity ID (meaning Entity Linking has failed to link them to IDs) are ignored in this loss. 

Ablation results (c.f. Table~\ref{table:ablations_coref}) shows that this auxiliary loss is not meaningfully important for model performance, although it does gives a modest boost to the ROUGE-L score on the \narrativeqa task.

\paragraph{MLM Analysis}
\begin{table}
\centering
\resizebox{\columnwidth}{!}{%
\begin{tabular}{lrr}
    \multirow{2}{*}{\textbf{Model}} & \multicolumn{2}{c}{\textbf{MLM acc (valid),\%}} \\
    & entity tokens & all tokens \\
    \hline
    \rit (SS) & 42.9 & 48.9 \\ 
    \rit & 50.3 & 50.5 \\     
\end{tabular}
}
\caption{\label{table:mlm} Masked Language Model (MLM) accuracy on the held out set. \textbf{SS} mode corresponds to the case when memories are collected only from the segment itself, effectively disabling any information propagation between segments.}
\end{table}
We consider whether the models learns to use memories in its predictions, evaluating MLM accuracy on a heldout set of Wikipedia articles and books. We compare \rit with a restricted version of itself that cannot access memories from different segments, denoted as \textit{Single Segment} (SS) setting. The single segment version of RIT is essentially a standard \roberta model with an additional attention mechanism over entity mentions within the segment, but different segments are processed completely independently. The results are reported in Table~\ref{table:mlm}. \rit achieves a $+1.5\%$ accuracy gain over \rit (SS), rising to $+7.4\%$ for entity tokens, confirming the model learns to utilize mention memory.

\begin{table}[t]
\centering
\small
\begin{tabular}{lcccc}
    \textbf{Model} & \textbf{\textsc{HQA}} & \textbf{\textsc{NQA}}-R & \textbf{\textsc{NQA}}-B & \textbf{\textsc{TQA}}\\
    \hline
    \rit $\mathtt{E}$ & 75.89 & 22.71 & 21.07 & 80.7\\
    \hline
    w/o coref & 76.0 & 21.79 & 21.01 & 80.5 \\
\end{tabular}
\caption{\label{table:ablations_coref} Ablation studies on variants of \rit. We report F1 (answer only) score for \textsc{HQA}, ROUGE-L and BLEU-1 for \textsc{NQA} (-R and -B correspondingly) and F1 for \textsc{TQA}.}
\end{table}

\section{Question Answering}

\begin{table}
\centering
\begin{tabular}{lrr}
    \textbf{Model} & \textbf{\#Params} \\
    \hline
    \bert \citep{bert} & 110M \\
    \roberta \citep{roberta} & 125M \\
    LF \citep{longformer} & 149M\\
    \etc \citep{etc} & 166M\\
    \bigbird \citep{bigbird} & 166M\\
    \hline
    \rit($\mathtt{ENTITY}$) & 145M
\end{tabular}
\caption{\label{table:model_params} Number of parameters per model.}
\end{table}

\begin{table*}
\centering
\begin{tabular}{lrr|rr|rr}
    \multirow{2}{*}{\textbf{Model}} & \multicolumn{2}{c}{\textbf{dev}} & 
    \multicolumn{2}{c}{\textbf{test (full)}} & 
    \multicolumn{2}{c}{\textbf{test (verified)}}\\
    & \textbf{F1} & \textbf{EM} & \textbf{F1} & \textbf{EM} & \textbf{F1} & \textbf{EM} \\
    \hline
    LF & 75.2 & - & - & - & - & - \\
    \bigbird & 79.5 & 75.7 & - & - & - & - \\
    \hline
    \roberta (us) & 75.9 & 71.3 & - & - & - & - \\
    \rit($\mathtt{ENTITY}$) & 80.7 & 76.47 & 80.9 & 76.7 & 89.2 & 86.8 \\ 
\end{tabular}
\caption{\label{table:trivia_qa_full} Results on the \triviaqa dataset in Wikipedia setting.}
\end{table*}

\subsection{Extractive QA layers}
The model is fine-tuned and evaluated on several extractive QA tasks. We introduce additional QA-specific layers to generate span-based predictions. Let $H_i$ be model's output for the segment $i$. The model generates separate scores (logits) for whether a token $j$ is the beginning of an answer span, $Z^{(b)}_{i,j}$, and another score for the end of the span, $Z^{(e)}_{i,j}$
\begin{align}
    Z^{(b)}_{i,j} &= W_b \cdot \mathtt{FFN}(H_{i,j})\\
    Z^{(e)}_{i,j} &= W_e \cdot \mathtt{FFN}(H_{i,j}) 
\end{align}
where $W_b, W_e$ are learnable weights and $\mathtt{FFN}(\cdot)$ is a shared fully-connected layer. 

For the loss function we largely follow works by \citep{ClarkGardner,ProbAssumptionsQA}, which describe an efficient way to train an extractive QA system in multi-segment setting where there are multiple correct answer spans in the evidence. Let $\mathcal{B} = \{(i, j) \mid \text{answer span starts at a position } j \text{ in the segment } i\}$. Then the loss is OR-model with global normalization
\begin{align}
    \label{eq:span_loss}
    \mathcal{L}_{span}^{(b)} &= -\log \frac{\sum_{(i, j) \in \mathcal{B}} \exp{Z^{(b)}_{i, j}}}{\sum_{i, j} \exp{Z^{(b)}_{i, j}}}
\end{align}
The loss for the end position of the answer span $\mathcal{L}_{span}^{(e)}$ is computed in a similar way. During inference, the model picks the most confident prediction amongst all the segments.

\subsection{\hotpotqa}
\textbf{Data download link}: \url{https://hotpotqa.github.io}

Unlike other datasets \hotpotqa contains questions with ``yes`` / ``no`` answers. In order to handle them appropriately we introduce an additional classification layer on top of \rit output CLS representation. The layer produces scores for all three possible options -- the answer is ``yes``, the answer is ``no`` and the answer is a span in the document. During training we normalize these scores globally across all passages. The loss function for the option classifier is a negative log-likelihood of the correct option, applied only to the two supporting paragraphs (not the distractors). During inference we select the option with the highest score across all paragraphs. If the selected option is ``yes`` or ``no``, we use it as the model's prediction. If the classifier predicts that answer is a span then we use the standard extractive QA layer to extract a span.

\textbf{Model selection} Model was selected based on the highest F1 (answer only) score on the development set. The best model was trained with a learning rate $3 \times 10^{-5}$ for 6 epochs.


\subsection{\triviaqa}
\textbf{Data Download Link}: \url{https://nlp.cs.washington.edu/triviaqa}

A complete set of \triviaqa evaluation results is shown in the Table~\ref{table:trivia_qa_full}.

\textbf{Model selection} Model was selected based on the highest F1 score on the development set. The best model was trained with a learning rate $1 \times 10^{-5}$ for 5.3 epochs.

\subsection{\narrativeqa}
\textbf{Data Download Link}: \url{https://github.com/deepmind/narrativeqa}

Here we provide more information on the NarrativeQA dataset. In particular, we would like to point that the manner in which NarrativeQA questions were generated encourages questions to require information from distant parts of the documents. 

Every question was written given a short Wikipedia summary of a movie/book as context. Accordingly, models can achieve high accuracy on NarrativeQA when given the summary as evidence, rather than the whole book (\citep{frhard} \citet{ProbAssumptionsQA} report ROUGE-L scores of 57.19 and 60.5 on the dev set, respectively). However, each sentence in the Wikipedia summary might correspond to a whole chapter of a book, so any question that uses information from multiple summary sentences is likely to require information from different sections of the book. Indeed, retrieve-and-read methods perform poorly in this setting~\citep{frhard}. On the other hand, \rit performs significantly better, demonstrating its ability to capture long-term dependencies. 

\paragraph{Preprocessing} In contrast to the \hotpotqa and \triviaqa, answers to NarrativeQA questions do not necessarily correspond to spans in the document. An exact answer can be found only in $\approx40\%$ cases - for the rest of the questions we use a ROUGE-L oracle as labels.  

\paragraph{Fine-tuning} Similarly to \citet{fevry2020eae} we found it helpful to enforce sparsity in the $\mathtt{MemoryAttention}$ layer by computing attention in the Equation 1 and 4 only over the 100 memories $m$ which have the largest dot product with the hidden state $h_{ij}$.

\paragraph{Evaluation} 
In line with the previous work \citep{narrativeqa, frhard} we convert both hypothesis and reference to lowercase and remove a trailing period before running an evaluation script. Following \citet{frhard} we use an open-source library to perform evaluation\footnote{\url{https://github.com/Maluuba/nlg-eval} by \citet{nlgeval}}, which includes ROUGE-L, BLEU-1, BLEU-4 and METEOR scores.

\textbf{Model selection} Model was selected based on the highest ROUGE-L score on the development set. The best model was trained with a learning rate $5 \times 10^{-6}$ for 2.2 epochs.

\end{document}